# Additive Pattern Database Heuristics

**Ariel Felner**                                                          FELNER@BGU.AC.IL
*Department of Information Systems Engineering,*
*Ben-Gurion University of the Negev, Beer-Sheva, 85104, Israel*

**Richard E. Korf**                                                       KORF@CS.UCLA.EDU
*Department of Computer Science, University of California, Los Angeles, 90095*

**Sarit Hanan**                                                           SARIT@ESHCOLOT.COM
*Department of Computer Science, Bar-Ilan University*
*Ramat-Gan, Israel, 52900*

## Abstract

We explore a method for computing admissible heuristic evaluation functions for search problems. It utilizes pattern databases (Culberson & Schaeffer, 1998), which are precomputed tables of the exact cost of solving various subproblems of an existing problem. Unlike standard pattern database heuristics, however, we partition our problems into disjoint subproblems, so that the costs of solving the different subproblems can be added together without overestimating the cost of solving the original problem. Previously (Korf & Felner, 2002) we showed how to statically partition the sliding-tile puzzles into disjoint groups of tiles to compute an admissible heuristic, using the same partition for each state and problem instance. Here we extend the method and show that it applies to other domains as well. We also present another method for additive heuristics which we call dynamically partitioned pattern databases. Here we partition the problem into disjoint subproblems for each state of the search dynamically. We discuss the pros and cons of each of these methods and apply both methods to three different problem domains: the sliding-tile puzzles, the 4-peg Towers of Hanoi problem, and finding an optimal vertex cover of a graph. We find that in some problem domains, static partitioning is most effective, while in others dynamic partitioning is a better choice. In each of these problem domains, either statically partitioned or dynamically partitioned pattern database heuristics are the best known heuristics for the problem.

## 1. Introduction and Overview

Heuristic search is a general problem-solving method in artificial intelligence. The most important heuristic search algorithms include A*(Hart, Nilsson, & Raphael, 1968), iterative-deepening-A* (IDA*)(Korf, 1985a), and depth-first branch-and-bound (DFBnB). All of these algorithms make use of a heuristic evaluation function $h(n)$, which takes a state $n$ and efficiently computes an estimate of the cost of an optimal solution from node $n$ to a goal state. If the heuristic function is "admissible", meaning that it never overestimates the cost of reaching a goal, then all the above algorithms are guaranteed to return an optimal solution, if one exists. The most effective way to improve the performance of a heuristic search algorithm is to improve the accuracy of the heuristic evaluation function. Developing more accurate admissible heuristic functions is the goal of this paper.

## 1.1 Sliding-Tile Puzzles

One of the primary examples in this paper is the well-known family of sliding-tile puzzles (see Figure 1). The Fifteen Puzzle consists of fifteen numbered tiles in a $4 \times 4$ square frame, with one empty position, called the "blank". A legal move is to move any tile horizontally or vertically adjacent to the blank into the blank position. The task is to rearrange the tiles from some initial configuration into a particular desired goal configuration. An optimal solution to the problem uses the fewest number of moves possible. The Twenty-Four Puzzle is the $5 \times 5$ version of this problem.

Figure 1: The Fifteen and Twenty-Four Puzzles in their Goal States

The classic evaluation function for the sliding-tile puzzles is called Manhattan distance. It is computed by counting the number of grid units that each tile is displaced from its goal position, and summing these values over all tiles, excluding the blank. Since each tile must move at least its Manhattan distance to its goal position, and a legal move only moves one physical tile, the Manhattan distance is a lower bound on the minimum number of moves needed to solve a problem instance.

Using the Manhattan distance heuristic, IDA* can find optimal solutions to randomly generated Fifteen puzzle instances in less than a minute on current machines. However, optimally solving random instances of the Twenty-Four puzzle with Manhattan distance is not practical on current machines. Thus, a more accurate heuristic function is needed.

## 1.2 Subgoal Interactions

The reason for the inaccuracy of the Manhattan distance function is that it assumes that each tile can be moved along a shortest path to its goal location, without interference from any other tiles. What makes the problem difficult is that the tiles do interfere with each other. The key to developing a more accurate heuristic function is to account for some of those interactions in the heuristic.

## 1.3 Overview of Paper

This paper is organized as follows. In section 2 we present previous work on designing more accurate heuristic functions, including pattern databases in general (Culberson & Schaeffer, 1998; Korf, 1997) and statically-partitioned additive pattern databases in particular (Korf & Felner, 2002).

In section 3 we present the idea of dynamically-partitioned additive pattern databases, initially in the context of the tile puzzles. This is another approach to additive pattern databases where the partitioning into disjoint subproblems is done dynamically for each state of the search, rather then statically in advance for all the states of the search (Korf & Felner, 2002).

Sections 4-6 present three applications of these ideas. Section 4 presents our implementations and experiments on the tile puzzles. We then present two other applications of additive pattern databases, including the 4-peg Towers of Hanoi problem in Section 5, and finding a minimum vertex cover of a graph in Section 6. We discuss the method in general in Section 7, and finally suggest further work and present our conclusions in Section 8.

The basic idea of dynamically-partitioned pattern databases was developed independently by Gasser (Gasser, 1995) and by Korf and Taylor (Korf & Taylor, 1996) in the context of the sliding-tile puzzles. It was also briefly mentioned in the earlier work(Korf & Felner, 2002) where statically-partitioned databases were introduced. Much of the material on previous work described below was taken from (Korf & Felner, 2002).

There are two main contributions of this paper. The first is that we show that additive pattern databases, which were first developed and implemented for the sliding-tile puzzles, can be generalized and applied to other domains. The second is that we divide additive pattern databases into two methods, statically- and dynamically-partitioned databases, and compare the two methods experimentally.

## 2. Previous Work on Admissible Heuristic Functions

In this section, we describe previous work on admissible heuristics from the classic theories until the new idea of pattern databases.

### 2.1 Heuristics as Optimal Solutions to Relaxed Problems

In general, admissible heuristic functions represent the cost of exact solutions to simplified or relaxed versions of the original problem (Pearl, 1984). For example, in a sliding-tile puzzle, to move a tile from position $x$ to position $y$, $x$ and $y$ must be adjacent, and position $y$ must be empty. By ignoring the empty constraint, we get a simplified problem where any tile can move to any adjacent position. We can solve any instance of this new problem optimally by moving each tile along a shortest path to its goal position, counting the number of moves made. The cost of such a solution is exactly the Manhattan distance from the initial state to the goal state. Since we removed a constraint on the moves, any solution to the original problem is also a solution to the simplified problem, and the cost of an optimal solution to the simplified problem is a lower bound on the cost of an optimal solution to the original problem. Thus, any heuristic derived in this way is admissible.

### 2.2 Linear Conflicts

The first significant improvement to Manhattan distance was the linear-conflict heuristic (Hansson, Mayer, & Yung, 1992). Assume that in the goal state, tile 1 is to the left of tile 2 in the top row, but in some particular state, tile 2 is to the left of tile 1 in the top row. The Manhattan distance of these tiles counts the number of steps they must move in the top row to reach their goal positions, but doesn't account for the fact that one of these tiles must move out of the top row to allow the other to pass by, and then move back into the top row. Thus, the interaction between these two tiles allows us to add two moves to the sum of their Manhattan distances without violating admissibility. The full linear-conflict heuristic finds all tiles that are in their goal row and/or column, but reversed in order, computes the number of moves needed to resolve these conflicts, and adds this to the Manhattan distance heuristic.

## 2.3 Non-Additive Pattern Databases

### 2.3.1 FIFTEEN PUZZLE

(Culberson & Schaeffer, 1998) carried this idea much further. For any given state, the minimum number of moves needed to get any subset of the tiles to their goal positions, including moves of other tiles, is clearly a lower bound on the number of moves needed to solve the entire puzzle. They chose as a subset of the Fifteen Puzzle tiles those in the bottom row and those in the rightmost column, referring to these as the *fringe* tiles.

The number of moves needed to solve the fringe tiles depends on the current positions of the fringe tiles and the blank, but is independent of the positions of the other tiles. We can precompute a table of these values, called a *pattern database*, with a single breadth-first search backward from the goal state. In this search, the unlabeled tiles are all equivalent, and a state is uniquely determined by the positions of the fringe tiles and the blank. As each configuration of these tiles is encountered for the first time, the number of moves made to reach it is stored in the corresponding entry of the table. Note that this table is only computed once for a given goal state, and the cost of computing it can be amortized over the solution of multiple problem instances with the same goal state.

Once this table is computed, IDA* can be used to search for an optimal solution to a particular problem instance. As each state is generated, the positions of the fringe tiles and the blank are used to compute an index into the pattern database, and the corresponding entry is used as the heuristic value of that state.

Using the fringe and another pattern database, and taking the maximum of the two database values as the overall heuristic value, Culberson and Schaeffer reduced the number of nodes generated to solve random Fifteen Puzzle instances by a factor of about a thousand, and reduced the running time by a factor of twelve, compared to Manhattan distance (Culberson & Schaeffer, 1998). Since these database values include moves of tiles that are not part of the pattern, they are *non-additive*, and the only way to admissibly combine the values from two such databases is to take their maximum value.

### 2.3.2 RUBIK'S CUBE

Non-additive pattern databases were also used to find the first optimal solutions to the $3 \times 3 \times 3$ Rubik's Cube (Korf, 1997). Three different pattern databases were precomputed. One stored the number of moves needed to solve the eight corner pieces, another contained the moves needed for six of the twelve edge pieces, and the other covered the remaining six edge pieces. Given a state in an IDA* search, we use the configurations of each of the three groups of pieces to compute indices into the corresponding pattern databases, and retrieve the resulting values. Given these three heuristic values, the best way to combine them without sacrificing admissibility is to take their maximum, since every twist of the cube moves four edge pieces and four corner pieces, and moves that contribute to the solution of pieces in one pattern database may also contribute to the solution of the others. Using this method, IDA* was able to find optimal solutions to random instances of Rubik's Cube (Korf, 1997). The median optimal solution length is only 18 moves. With improvements by Michael Reid and Herbert Kociemba, larger pattern databases, and faster computers, most states can now be solved in a matter of hours.

### 2.3.3 Limitations of Non-Additive Pattern Databases

One of the limitations of non-additive pattern databases is that they don't scale up to larger problems. For example, since the Twenty-Four puzzle contains 25 different positions, a pattern database covering $n$ tiles and the blank would require $25!/(25 - n - 1)!$ entries. A database of only six tiles and the blank would require over 2.4 billion entries. Furthermore, the values from a database of only six tiles would be smaller than the Manhattan distance of all the tiles for almost all states.

If we divide the tiles into several disjoint groups, the best way to combine them admissibly, given the above formalization, is to take the maximum of their values. The reason is that non-additive pattern database values include all moves needed to solve the pattern tiles, including moves of other tiles.

Instead of taking the *maximum* of different pattern database values, we would like to be able to *sum* their values, to get a more accurate heuristic, without violating admissibility. We present two ways to do this: statically-partitioned additive pattern databases, and dynamically-partitioned additive pattern databases. In this paper we study both these methods, present general conditions for their applicability, and compare them experimentally.

## 2.4 Statically-Partitioned Additive Database Heuristics

When statically-partitioned additive pattern databases were introduced (Korf & Felner, 2002), they were called *disjoint pattern databases*. We introduce the new terminology here to more clearly distinguish them from dynamically-partitioned pattern databases that will be presented below. To construct a statically-partitioned pattern database for the sliding-tile puzzles, we partition the tiles into disjoint groups, such that every tile is included in a group, and no tile belongs to more than one group. We precompute pattern databases of the minimum number of moves of the tiles in each group that are required to get those tiles to their goal positions. Then, given a particular state in the search, for each group of tiles, we use the positions of those tiles to compute an index into the corresponding pattern database, retrieve the number of moves required to solve the tiles in that group, and then *add* together the values for each group, to compute an overall admissible heuristic for the given state. This value will be at least as large as the Manhattan distance of the state, and usually larger, since it accounts for interactions between tiles in the same group. The term "statically-partitioned" refers to the fact that the same partition of the tiles is used for all states of the search.

The key difference between additive databases and the non-additive databases described in section 2.3 above is that the non-additive databases include all moves required to solve the pattern tiles, including moves of tiles not in the pattern group. As a result, given two such database values, even if there is no overlap among their tiles, we can only take the maximum of the two values as an admissible heuristic, because moves counted in one database may move tiles in the other database, and hence these moves would be counted twice. In an additive pattern database, we only count moves of the tiles in the group.

Manhattan distance is a trivial example of a set of additive pattern databases, where each group contains only a single tile. For each tile, we could use the breadth-first search described above to automatically compute a table of the minimum number of moves of that tile needed to move it from any position to its goal position. Such a set of tables would contain the Manhattan distance of each tile from each position to its goal position. Then, given a particular state, we simply look up the position of each tile in its corresponding table and sum the resulting values, thus computing the sum of the Manhattan distances. In fact, an efficient implementation of Manhattan distance works in exactly this way, looking up the position of each tile in a precomputed table of Manhattan distances, rather than computing it from $x$ and $y$ coordinates of the current and goal positions of the tile.

In the earlier paper (Korf & Felner, 2002) we implemented statically-partitioned additive pattern databases for both the Fifteen Puzzle and the Twenty-Four puzzles. Define an $x - y - z$ partitioning to be a partition of the tiles into disjoint sets with cardinalities of $x$, $y$ and $z$. For the Fifteen Puzzle we used a 7-8 partitioning and for the Twenty-Four Puzzle we used a 6-6-6-6 partitioning. Currently, these implementations are the best existing optimal solvers for these puzzles.

### 2.4.1 LIMITATIONS OF STATICALLY-PARTITIONED DATABASE HEURISTICS

The main limitation of the statically-partitioned pattern database heuristics is that they fail to capture interactions between tiles in different groups of the partition. In this paper we will try to address this limitation by adding a different approach to additive pattern databases, namely dynamically-partitioned database heuristics. Throughout the paper we will use both approaches for the various domains, and study the pros and cons of each.

## 3. Dynamically-Partitioned Database Heuristics

The main idea behind this section was developed independently by (Gasser, 1995) and (Korf & Taylor, 1996), in the context of the sliding-tile puzzles. Consider a table which contains for each pair of tiles, and each possible pair of positions they could occupy, the number of moves required of those two tiles to move them to their positions. Gasser refers to this table as the *2-tile pattern database*, while we call these values the *pairwise distances*. For most pairs of tiles in most positions, their pairwise distance will equal the sum of their Manhattan distances. For some tiles in some positions however, such as two tiles in a linear conflict, their pairwise distance will exceed the sum of their Manhattan distances. Given $n$ tiles, there are $O(n^4)$ entries in the complete pairwise distance table, but only those pairwise distances that exceed the sum of the Manhattan distances of the two tiles need be stored. For example, the full pairwise distance table for the Twenty-Four puzzle would contain $(24 \cdot 23/2) \cdot 25 \cdot 24 = 165,600$ entries, but only about 3000 of these exceed the sum of their Manhattan distances.

## 3.1 Computing the Pairwise Distances

How do we compute such tables? For each pair of tiles, we perform a single breadth-first search, starting from the goal state. In this search, a state is uniquely determined by the positions of the two tiles of interest, and the blank. All other tiles are indistinguishable. The first time that each different state of the two tiles is reached, regardless of the position of the blank, the number of moves made to reach this state is recorded in the database. The search continues until all states of the two tiles and the blank have been generated.

We perform a separate search for each of the $n(n-1)/2$ different pairs of tiles, where $n$ is the number of tiles. Since each state of these searches is determined by the positions of the two tiles and the blank, there are $O(n^3)$ different states in each search, for an overall complexity of $O(n^5)$ to compute all the pairwise-distance tables. Note that these tables are only computed once for a given goal state, so their computational cost can be amortized over all subsequent problem-solving trials with the same goal state.

## 3.2 Computing the Heuristic Value of a State

Given a 2-tile database, and a particular state of the puzzle, we can't simply sum the database values for each pair of tiles to compute the heuristic, since each tile is paired with every other tile, and this sum will grossly overestimate the optimal solution length. Rather, we must partition the $n$ tiles into $n/2$ non-overlapping pairs, and then sum the pairwise distances for each of the chosen pairs. With an odd number of tiles, one tile will be left over, and simply contributes its Manhattan distance to the sum. To get the most accurate admissible heuristic, we want a partition that maximizes the sum of the pairwise distances. For each state of the search, this maximizing partition may be different, requiring the partitioning to be performed for each heuristic evaluation. Thus, we use the term *dynamically-partitioned additive pattern database heuristics*.

To compute the heuristic for a given state, define a graph where each tile is represented by a vertex, and there is an edge between each pair of vertices, labeled with the pairwise distance of the corresponding pair of tiles in that state. We call this graph the *mutual-cost graph*. The task is to choose a set of edges from this graph so that no two chosen edges are incident to the same vertex, such that the sum of the labels of the chosen edges is maximized. This is called the *maximum weighted matching* problem, and can be solved in $O(n^3)$ time (Papadimitriou & Steiglitz, 1982), where $n$ is the number of vertices, or tiles in this case.

## 3.3 Triple and Higher-Order Distances

This idea can generalized to larger groups of tiles as follows. Let $k$ be the size of the group. A $k-tile$ database includes a table for each of the $\binom{n}{k}$ different groups of $k$ tiles. Each table includes the number of moves of these $k$ tiles that are required to get them to their goal positions from each possible set of $k$ positions they could occupy. A degenerate case is a set of $1-tile$ databases which is actually a lookup table with the Manhattan distances. The next case is the pairs database described above. Potentially there are $(n+1)n(n-1)(n-2)\ldots(n-k+2)$ different entries in each table, since $n$ is the number of tiles (such as 15) while $n+1$ is the number of positions (such as 16). However while building a set of $k - tile$ databases we don't necessarily need all these values if we also have the set of the $(k - 1) - tile$ databases. For example, suppose that we are creating a set of the $2 - tile$ databases and that we already have all the Manhattan distances (1-tile databases). In that case we only need to store pairwise distances that are greater than the sum of the Manhattan distances of the two tiles in the pairs. In general in a $k - tile$ database we only need to store a value that exceeds the sum of any of the partitions of the $k$ tiles to smaller size groups.

The next step is to have $3 - tile$ databases. A set of $3 - tile$ databases contains, for each triple of tiles and each possible set of three positions they could occupy, the number of moves of the three tiles required to get them to their goal positions. We only need to store those values that exceed the both the sum of the Manhattan distances, as well as the value given to these three tiles by the pairwise heuristic of the corresponding tiles. With the addition of the $3 - tile$ databases, the corresponding mutual-cost graph contains a vertex for each tile, an edge for each pairwise distance, and a hyperedge connecting three vertices for each triple distance. The task is to choose a set of edges and hyperedges that have no vertices in common, so that the sum of the weights of the edges and hyperedges is maximized. Unfortunately, the corresponding three-dimensional matching problem is NP-complete (Garey & Johnson, 1979), as is higher-order matching. For the tile puzzles, however, if we only include tiles whose pairwise or triple distances exceed the sum of their Manhattan distances, the mutual-cost graph becomes very sparse, and the corresponding matching problem can be solved relatively efficiently, as presented below. For problems where the mutual-cost graph is not sparse, then we might not be able to optimally solve the maximum-matching problem and will have to settle for a sub optimal matching which is still an admissible heuristic to the original problem. For example, this is the case with the vertex-cover domain of section 6.

The main advantage of this approach is that it can capture more tile interactions, compared to a statically-partitioned pattern database. The disadvantage of this approach is that computing the heuristic value of each state requires solving a matching problem on the mutual-cost graph, which is much more expensive than simply adding the values from a database for each group.

We now consider these ideas for each of our problem domains in turn: sliding-tile puzzles, the 4-Peg Towers of Hanoi problem, and vertex cover. We compare static and dynamic partitioning for each of these domains.

## 4. Sliding-Tile Puzzles

In this section we describe our experiments solving the Fifteen and Twenty-Four puzzles using statically- and dynamically-partitioned additive pattern database heuristics. We begin by describing a domain-specific enhancement for computing an admissible heuristic for the sliding-tile puzzles for the dynamically-partitioned pattern databases which is more accurate than the general maximum weighted matching technique described above.

### 4.1 Weighted Vertex Cover Heuristic

Consider a state of a sliding-tile puzzle in which three tiles are in their goal row, but in opposite order from their goal order. Each of these tiles is in a linear conflict with the other two. We can represent this 3-way linear conflict with a triangle graph, shown in Figure 2, where each tile is represented by a vertex, and there is an edge between each pair of vertices. In the mutual-cost graph, the label on each edge would be the sum of the Manhattan distances of the two tiles represented by the vertices at each end of the edge, plus two moves to resolve the linear conflict between them. To simplify the problem, we subtract the Manhattan distances from the edges, resulting in the *conflict graph* shown in Figure 2. What is the largest admissible heuristic for this situation?

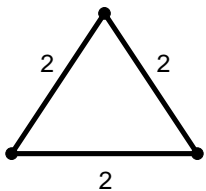

Figure 2: Conflict graph for 3-way linear conflict

The maximum matching of this graph can only contain one edge, with a cost of two, since any two edges will have a vertex in common. However, in this case we can actually add four moves to the sum of the Manhattan distances of the three tiles, because two of the three tiles will have to temporarily move out of the goal row. Thus, while the maximum weighted matching is clearly admissible, it doesn't always yield the largest possible admissible heuristic value. If the pairwise distance of tiles $X$ and $Y$ in a given state is $a$, there will be an edge in the corresponding mutual-cost graph between nodes $X$ and $Y$, weighted by $a$. If $x$ is the number of moves of tile $X$ in a solution, and $y$ is the number of moves of tile $Y$, then their pairwise distance represents a constraint that $x + y \geq a$. Each edge of the mutual-cost graph represents a similar constraint on any solution. For simplicity and efficiency, we assume here that the Manhattan distance of each tile will be added to the final heuristic value, so the conflict graph will only represent moves in addition to the Manhattan distance of each tile.

The problem is to assign a number of moves to each vertex of the conflict graph such that all the pairwise distance constraints are satisfied. Since the constraints are all lower bounds, assigning large values to each node will satisfy all the constraints. In order for the resulting heuristic to be admissible, however, the sum of the values of all vertices must be the minimum sum that satisfies all the constraints. This sum is then the maximum admissible heuristic for the given state. Furthermore, each vertex must be assigned a non-negative integer. In the case of the triangle conflict graph in Figure 2, assigning a value of one to each vertex or tile, for an overall heuristic of three, is the minimum value that will satisfy all the pairwise constraints.

While this is true for the general case, we can take it further for the tile puzzle domain. For the special case of the tile puzzle, any path between any two locations must have the same even-odd parity as the Manhattan distance between the two locations. Therefore if the pairwise distance of X and Y is larger than the sum of their Manhattan distance by two, then at least one of the tiles must move at least two moves more than its Manhattan distance in order to satisfy the pairwise conflict. Therefore, such an edge in the pairwise conflict graph means not only that $x + y \geq 2$ but that $x \geq 2$ or $y \geq 2$.

In order to satisfy all these constraints, one of the vertices incident to each edge must be set to at least two. If all edges included in the conflict graph have cost two, then the minimal assignment is two times the number of vertices needed such that each edge is incident to one of these vertices. Such a set of vertices is called a *vertex cover*, and the smallest such set of vertices is called a *minimum vertex cover*. A minimum vertex cover of the graph in Figure 2 must include two vertices, for an overall heuristic of four, plus the sum of the Manhattan distances.

The vertex-cover heuristic is easily generalized from pairwise edges to handle triples of nodes. For example, each triple distance that is not captured by Manhattan or pairwise distances introduces a hypergraph "edge" that connects three nodes. The corresponding constraint is that the sum of the costs assigned to each of the three endpoints must be greater than or equal to the weight of the hyperedge. For the vertex cover, a hyperedge is covered if any of its three endpoints are in the set.

There are two possibilities to cover a pairwise edge $(X, Y)$ with a weight of two i.e., assigning two moves either to $X$ or to $Y$. However, some triple distances are four moves greater than the sum of their Manhattan distances. Covering hyperedges with weights larger than two is a little more complicated but the principle is the same and all possibilities are considered. See (Felner, 2001) for more details.

The general problem here can be called "weighted vertex cover". Given a hypergraph with integer-weighted edges, assign an integer value to each vertex, such that for each hyperedge, the sum of the values assigned to each vertex incident to the hyperedge is at least as large as the weight of the hyperedge. The "minimum weighted vertex cover" is a weighted vertex cover for which the sum of the vertex values is the lowest possible.

It is easy to prove that the minimum weighted vertex cover problem is NP-complete. Clearly it is in NP, since we can check a possible assignment of values to the vertices in time linear in the number of hyperedges. We can prove it is NP-complete by reducing standard vertex cover to it. Given a graph, the standard vertex-cover problem is to find a smallest set of vertices such that every edge in graph is incident to at least one vertex in the set. Given a standard vertex-cover problem, create a corresponding weighted vertex-cover problem by assigning a weight of one to every edge. The solution to this weighted vertex-cover problem will assign either zero or one to each vertex, such that at least one endpoint of each edge is assigned the value one. The original vertex-cover problem will have a solution of $k$ vertices if and only if there is a solution to the weighted vertex-cover problem in which $k$ vertices are assigned the value one.

Since weighted vertex cover is NP-complete, and we have to solve this problem to compute the heuristic for each state of the search, it may seem that this is not a practical approach to computing heuristics. Our experimental results show that it is, for several reasons. The first is that by computing the Manhattan distance first, and only including those edges for which the distance exceeds the sum of the Manhattan distances, the resulting conflict graph is extremely sparse. For a random state of the Twenty-Four Puzzle, it includes an average of only four pairs of nodes, and four triples of nodes, some of which may overlap. Secondly, since the graph is sparse, it is likely to be composed of two or more disconnected components. To find the minimum vertex cover of a disconnected graph, we simply add together the vertices in the minimum vertex covers of each component. Finally, in the course of the search, we can compute the heuristic incrementally. Given the heuristic value of a parent node, we compute the heuristic value of each child by focusing only on the difference between the parent and child nodes. In our case, only one tile moves between a parent and child node, so we incrementally compute the vertex cover based on this small change to the conflict graph.

## 4.2 Experimental Results

Here we compare our implementation of statically- and dynamically- partitioned additive pattern databases for both Fifteen and Twenty-Four puzzles. Note that the best statically-partitioned pattern databases were already reported in the earlier paper (Korf & Felner, 2002) and are provided here for purposes of comparison.

### 4.2.1 Mapping Functions

For the sliding-tile puzzles, there are two ways to store and index partial states in a pattern database, a sparse mapping and a compact mapping. For a pattern of $k$ tiles, a sparse table occupies a $k$-dimensional array with a total of $16^k$ entries for the Fifteen Puzzle. An individual permutation is mapped into the table by taking the location of each tile of the permutation as a separate index into the array. For example, if we have a pattern of three tiles (X,Y,Z), which are located in positions (2,1,3) they would be mapped to the array element a[2][1][3]. While this makes the mapping function efficient, it is inefficient in terms of space, since it wastes those locations with two or more equal indices.

Alternatively, a compact mapping occupies a single-dimensional array. A pattern of three tiles of the Fifteen puzzle needs an array of size $16 \times 15 \times 14$. Each permutation is mapped to an index corresponding to its position in a lexicographic ordering of all permutations. For example, the permutation (2 1 3) is mapped to the third index, since it is preceded by (1 2 3) and (1 3 2) in lexicographic order. This compact mapping doesn't waste space, but computing the indices is more complex and therefore consumes more time.

Note that in all cases below, except the 7-8 partitioning for the Fifteen Puzzle, a sparse mapping was used for the databases. For the 7-8 partitioning of the Fifteen Puzzle, we used a compact mapping. This was done to save memory as a sparse mapping for a pattern of 8 tiles exceeds our memory capacity.

### 4.2.2 Fifteen Puzzle

In section 2.4 under previous work, we mentioned a 7-8 statically-partitioned additive pattern database for the Fifteen Puzzle. Here we compare that static partition with several others, and with dynamically-partitioned additive pattern databases in order to study the behaviors of the different methods (Korf & Felner, 2002)[1].

---

1. It is worth mentioning here that the way we treated the blank tile for the statically-partitioned pattern databases is not trivial. For the backwards breadth-first search we treated this tile as a distinct tile that can move to any adjacent location. If that location was occupied by a real tile then that real tile was moved to the former blank location and we add one to the length of the path to that node. However, for the pattern database tables we have only considered the locations of the real tiles as indexes into the tables. We did not preserve the location of the blank and stored the minimum among all possible blank locations in order to save memory. In a sense, we have compressed the pattern databases according to the location of the blank (see (Felner, Meshulam, Holte, & Korf, 2004) about compressing pattern databases).

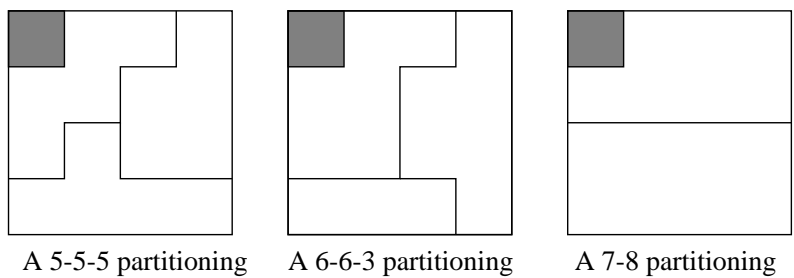

Figure 3: Different Statically-partitioned Databases for Fifteen Puzzle

| Heuristic Function | Value | Nodes | Sec. | Nodes/sec | Memory |
|---|---|---|---|---|---|
| Manhattan | 36.940 | 401,189,630 | 53 | 7,509,527 | 0 |
| Linear conflict | 38.788 | 40,224,625 | 10 | 3,891,701 | 0 |
| Dynamic, MM: pairs | 39.411 | 21,211,091 | 13 | 1,581,848 | 1,000 |
| Dynamic, MM: pairs+triples | 41.801 | 2,877,328 | 8 | 351,173 | 2,300 |
| Dynamic, WVC: pairs | 40.432 | 9,983,886 | 10 | 959,896 | 1,000 |
| Dynamic, WVC: pairs+triples | 42.792 | 707,476 | 5 | 139,376 | 2,300 |
| DWVC: pairs+triples+quadruples | 43.990 | 110,394 | 9 | 11,901 | 78,800 |
| Static: 5-5-5 | 41.560 | 3,090,405 | .540 | 5,722,922 | 3,145 |
| Static: 6-6-3 | 42.924 | 617,555 | .163 | 3,788,680 | 33,554 |
| Static: 7-8 | 45.630 | 36,710 | .028 | 1,377,630 | 576,575 |

Table 1: Experimental results on the Fifteen Puzzle.

Table 1 presents the results of running IDA* with different heuristics averaged over 1000 random solvable instances of the Fifteen Puzzle. The average optimal solution length was 52.552 moves. Each row presents results of a different heuristic.

The first column indicates the heuristic function. The second column, **Value** shows the average heuristic value of the thousand initial states. The **Nodes** column shows the average number of nodes generated to find an optimal solution. The **Seconds** column gives the average amount of CPU time that was needed to solve a problem on a 500 megahertz PC. The next column presents the speed of the algorithm in nodes per second. Finally, the last column shows the amount of memory in kilobytes that was needed for the databases, if any [2].

The first row presents the results of running Manhattan distance. The second row adds linear conflicts to the Manhattan distance. The next two rows present the results of the dynamically partitioned additive databases when the heuristic was obtained by a maximum-matching (MM) on the conflict graph. The next three rows present the results of the dynamically partitioned additive databases with the weighted vertex-cover heuristic (WVC), which is more accurate than the maximum-matching heuristic for this problem. Thus, we include the maximum-matching results for the Fifteen Puzzle for comparison

---

2. For simplicity, in this paper we assume that a kilobyte has $10^3$ bytes and that a megabyte has $10^6$ bytes

purposes, while for the Twenty-Four and Thirty-Five Puzzles below, we only used the weighted vertex-cover heuristics. We can view the weighted vertex-cover heuristic as an implementation of the dynamically partitioned database in this domain, since for each node of the search tree, we retrieve all the pairwise, triple distances etc, and find the best way to combine them in an admissible heuristic.

By only storing those pairs of positions where the heuristic value exceeds the sum of the Manhattan distances, the pairs databases only occupied about one megabyte of memory. By only storing those triples of positions that exceeded the sum of the Manhattan distances, and also were not covered by the pairs database, the pairs and triples together only occupied about 2.3 megabytes of memory. IDA* with the pairs heuristic (with weighted vertex-cover) required an average of about ten seconds to solve each instance, and the pairs and triples heuristic reduced this to about five seconds each. We have also experimented with a dynamically partitioned heuristic that also considers quadruples of tiles. However, while adding quadruples reduced the number of generated nodes, the constant time per node increased, and the overall running time was longer than the heuristic that only used pairs and triples. We have found that the bottleneck here was keeping the conflict graph accurate for each new node of the search tree. For every tile that moves we need to check all the pairs, triples and quadruples that include this tile to see whether edges were added or deleted from the conflict graph. This was expensive in time, and thus the system with pairs and triples produced the best results among the dynamically-partitioned heuristics.

For the statically-partitioned database we report results for three different partitions shown in Figure 3. The first uses a 5-5-5 partitioning of the tiles. The second uses a 6-6-3 partitioning. Finally, the third partition is the same 7-8 partitioning from (Korf & Felner, 2002). As the partitions use larger patterns, the memory requirements increase. While the 5-5-5 partitioning needed only 3 megabytes, the 7-8 partitioning needed 575 megabytes. The results for all the statically-partitioned databases were calculated by taking the maximum of the above partitions and the same partitions reflected about the main diagonal. The same tables were used for the reflected databases, simply by mapping the positions of the tiles to their reflected positions. Note that even the 5-5-5 partitioning is faster than the dynamically partitioned database while using the same amount of memory. The best results were obtained by the 7-8 partitioning and were already reported earlier (Korf & Felner, 2002). This heuristic reduced the average running time of IDA* on a thousand random solvable instance to 28 milliseconds per problem instance. This is over 2000 times faster than the 53 seconds required on average with the Manhattan distance heuristic on the same machine.

Thus, we conclude that if sufficient memory is available, heuristics based on statically-partitioned pattern databases are more effective for the Fifteen puzzle than those based on dynamically-partitioned pattern databases.

### 4.2.3 TWENTY-FOUR PUZZLE

In our earlier paper (Korf & Felner, 2002) we implemented a statically-partitioned additive pattern database for the Twenty-Four puzzle, partitioning the tiles into four groups of six tiles, as shown in Figure 4, along with its reflection about the main diagonal.

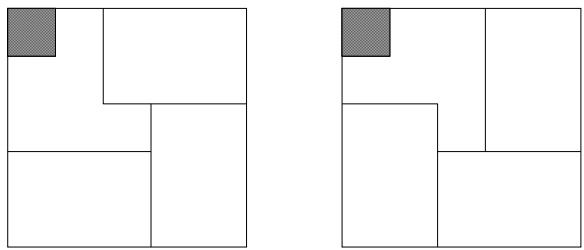

Figure 4: Static 6-6-6-6 database for Twenty-Four Puzzle and its reflection about the main diagonal

| Problem | | Dynamic Partitioning | | Static Partitioning | |
|---|---|---|---|---|---|
| No | Path | Nodes | Seconds | Nodes | Seconds |
| 1 | 95 | 306,958,148 | 1,757 | 2,031,102,635 | 1,446 |
| 2 | 96 | 65,125,210,009 | 692,829 | 211,884,984,525 | 147,493 |
| 3 | 97 | 52,906,797,645 | 524,603 | 21,148,144,928 | 14,972 |
| 4 | 98 | 8,465,759,895 | 72,911 | 10,991,471,966 | 7,809 |
| 5 | 100 | 715,535,336 | 3,922 | 2,899,007,625 | 2,024 |
| 6 | 101 | 10,415,838,041 | 151,083 | 103,460,814,368 | 74,100 |
| 7 | 104 | 46,196,984,340 | 717,454 | 106,321,592,792 | 76,522 |
| 8 | 108 | 15,377,764,962 | 82,180 | 116,202,273,788 | 81,643 |
| 9 | 113 | 135,129,533,132 | 747,443 | 1,818,055,616,606 | 3,831,042 |
| 10 | 114 | 726,455,970,727 | 4,214,591 | 1,519,052,821,943 | 3,320,098 |

Table 2: Twenty-Four Puzzle results for static vs. dynamic databases

Here we added experiments on the Twenty-Four Puzzle, using dynamically-partitioned pairwise and triple distances, and computing the heuristic as the minimum weighted vertex cover. These databases required about three megabytes of memory. On our 500 Megahertz Pentium III PC, IDA* using this heuristic generated about 185,000 nodes per second. We optimally solved the first ten randomly-generated problem instances from (Korf & Felner, 2002), and compared the results to those with the statically-partitioned pattern database heuristic. This heuristic generated 1,306,000 nodes per second.

Table 2 shows the results. The first column gives the problem number, and the second the optimal solution length for that problem instance. The next two columns show the number of nodes generated and the running time in seconds for IDA* with the dynamically-partitioned pattern database heuristic, and the last two columns present the same information for IDA* with the statically-partitioned database heuristic.

For the Twenty-Four puzzle, the dynamically-partitioned database heuristic usually results in fewer node generations, but a longer runtime overall, compared to the statically-partitioned database heuristic. The reason is that computing the best partition for each state, which requires solving a weighted vertex-cover problem, is much slower than simply summing values stored in the static pattern database. On the other hand, the dynamic database requires only three megabytes of memory, compared to 244 megabytes for the static database we used. Problem 3 in the table is the only case where the static partition generated a smaller number of nodes than the dynamic partition. Problem 9 is the only case where IDA* ran faster with the dynamically-partitioned heuristic than the statically-partitioned heuristic.

Here again we added quadruples to the dynamically-partitioned database, and found out that while the number of generated nodes decreased in all problems except problem 9, the overall running time increased, compared to using just pairwise and triple distances.

### 4.2.4 Thirty-Five Puzzle

Since the performance of the dynamic database relative to the static database improved in going from the Fifteen Puzzle to the Twenty-Four Puzzle, an obvious question is what happens with an even larger puzzle, such as the $6 \times 6$ Thirty-Five Puzzle. Unfortunately, finding optimal solutions to random instances of the Thirty-Five Puzzle is beyond the reach of current techniques and machines.

However, we can predict the relative performance of IDA* with different heuristics, using the new theory that was introduced lately (Korf, Reid, & Edelkamp, 2001). All that is needed is the brute-force branching factor of the problem space, and a large random sample of solvable states, along with their heuristic values. A rather surprising result of this theory is that the rate of growth of the number of nodes expanded by IDA* with increasing depth is exactly the brute-force branching factor. Different heuristics affect the exponent in this exponential growth, but not the base. To predict the actual performance of IDA* with a given heuristic, we would need the average optimal solution depth as well, which is unknown for the Thirty-Five Puzzle. However, we can predict the relative performance of different heuristics without knowing the solution depth. See (Korf et al., 2001) for more details.

We computed a dynamically-partitioned pairs and triples pattern database for the Thirty-Five Puzzle, storing only those values that exceed the sum of the Manhattan distances of the corresponding tiles. For each state, the heuristic is computed as the minimum weighted vertex cover of the conflict graph.

We also computed several different statically-partitioned pattern database heuristics. They were all based on a partitioning of the problem into 7 pattern of 5 tiles each. We ran them on random problems to limited depths, and compared the number of nodes generated to those depths. The one that generated the fewest nodes in searching to the same depth is shown in Figure 5. The two diagrams are reflections of each other about the main diagonal. Each group includes five tiles, and hence the number of entries in each database is $36!/31! = 45,239,040$. For speed of access, we stored each database in a table of size $36^5 = 60,466,176$. For the overall heuristic we used the maximum of the sum of the values from the seven databases, and the sum of the values from the seven databases reflected about the main diagonal.

In order to approximate the distribution of values of each heuristic, we sampled ten billion random solvable states of the problem, and computed both static and dynamically-partitioned heuristic values for each, as well as Manhattan distance. This information, along with the brute-force branching factor of the problem, which is approximately 2.36761 (Korf et al., 2001), allows us to predict the relative average number of nodes that would be generated by IDA* using each heuristic, when searching to the same depth. We computed the speed of IDA* using each heuristic, in nodes per second by running the algorithms to a limited depth. Table 3 below shows the results.

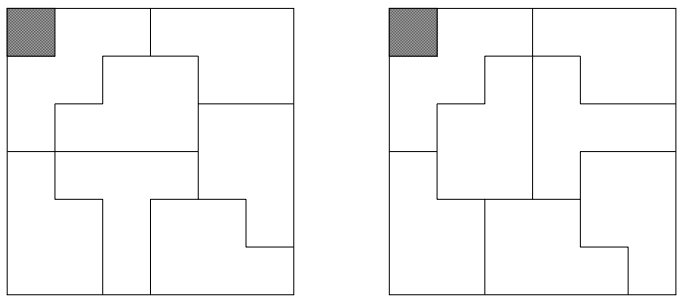

Figure 5: Disjoint Databases for Thirty-Five Puzzle

The first column gives the heuristic used with IDA*. The second column gives the average value of the heuristic over all ten billion random states. As expected, the dynamically partitioned heuristic has the highest average value, and Manhattan distance the lowest. The third column gives the number of nodes generated in searching to the same depth, relative to the dynamically-partitioned heuristic, which generates the fewest nodes. As expected, Manhattan distance generates the most nodes, and the dynamically-partitioned heuristic the fewest. The fourth column gives the speed in nodes per second of IDA* using each of the different heuristics on a 1.7 megahertz PC. As expected, Manhattan distance is the fastest, and the dynamically-partitioned heuristic the slowest. The fifth column gives the predicted relative running time of IDA* with each heuristic, searching to the same depth. This is computed by dividing the relative nodes by the nodes per second. The results are expressed relative to the dynamically predicted heuristic, which is the fastest overall. In this case, the slower speed in nodes per second of the dynamically-partitioned heuristic is more than made up for by the fewer node generations, making it about 1.8 times faster than the statically-partitioned heuristic overall. By comparison, IDA* with Manhattan distance is predicted to be almost 1000 times slower. The last column shows the memory required for the different heuristics, in kilobytes.

| Heuristic | Value | Rel. Nodes | Nodes/sec | Rel. Time | Memory |
|-----------|-------|-----------|-----------|-----------|--------|
| Manhattan | 135.02 | 31,000.0 | 20,500,000 | 987.5 | 0 |
| Static | 139.82 | 11.5 | 4,138,000 | 1.8 | 404,000 |
| Dynamic | 142.66 | 1.0 | 653,000 | 1.0 | 5,000 |

Table 3: Predicted Performance results on the Thirty-Five Puzzle

### 4.2.5 DISCUSSION OF TILE PUZZLE RESULTS

The relative advantage of the statically-partitioned database heuristics over the dynamically-partitioned heuristics appears to decrease as the problem size increases. When moving to the Thirty-Five Puzzle, the predicted performance of the dynamically partitioned database is better than the statically-partitioned database. One possible explanation is that for the Fifteen Puzzle, we can build and store databases of half the tiles. For the Twenty-Four Puzzle, we can only include a quarter of the tiles in a single database. For the Thirty-Five Puzzle, this decreases to only 1/7 of the tiles in one database. Note that for the Thirty-Five Puzzle, the memory needed by the dynamically-partitioned database is almost two orders of magnitude less than that required by the statically-partitioned database.

## 5. 4-Peg Towers of Hanoi Problem

We now turn our attention to another domain, the Towers of Hanoi problem.

### 5.1 Problem Domain

The well-known 3-peg Towers of Hanoi problem consists of three pegs that hold a number of different-sized disks. Initially, the disks are all stacked on one peg, in decreasing order of size. The task is to transfer all the disks to another peg. The constraints are that only the top disk on any peg can be moved at any time, and that a larger disk can never be placed on top of a smaller disk. For the 3-peg problem, there is a simple deterministic algorithm that provably returns an optimal solution. The minimum number of moves is $2^n - 1$ where $n$ is the number of disks.

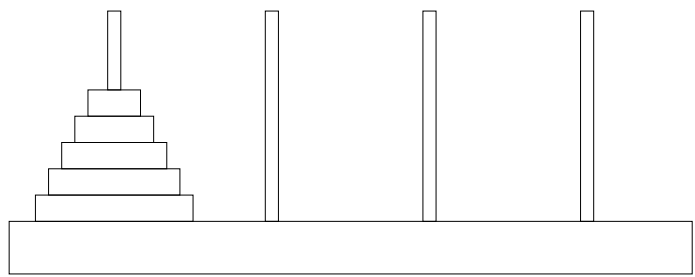

Figure 6: Five-disk four-peg Towers of Hanoi problem

The 4-peg Towers of Hanoi problem, known as Reve's puzzle (van de Liefvoort, 1992; Hinz, 1997) and shown in Figure 6, is more interesting. There exists a deterministic algorithm for finding a solution, and a conjecture that it generates an optimal solution, but the conjecture remains unproven (Frame, 1941; Stewart, 1941; Dunkel, 1941). Thus, systematic search is currently the only method guaranteed to find optimal solutions, or to verify the conjecture for problems with a given number of disks.

### 5.2 Search Algorithms

Once we eliminate the inverse of the last move made, the search tree for the sliding-tile puzzles generates relatively few duplicate nodes representing the same state. As a result, the depth-first algorithm iterative-deepening-A* (IDA*) is the algorithm of choice. This is not the case with the Towers of Hanoi problem, however.

### 5.2.1 BRANCHING FACTOR OF 4-PEG TOWERS OF HANOI

For the 4-peg problem, there are between three and six legal moves in any given state, depending on how many pegs are occupied. For example, if all pegs hold at least one disk, the smallest disk can move to any of the other three pegs, the next smallest disk that is on top of a peg can move to any of the two pegs that don't contain the smallest disk, and the third smallest disk that is on top of a peg can move to the single peg with a larger disk on top. We can reduce the number of duplicate nodes by never moving the same disk twice in a row, reducing the branching factor to between two and five. The average branching factor among those states with all pegs occupied, which is the vast majority of the states, is approximately 3.7664.

### 5.2.2 DEPTH-FIRST VS. BREADTH-FIRST SEARCH AND A*

Unfortunately, even with this reduction, there are many small cycles in the 4-peg Towers of Hanoi domain, meaning there are many different paths between the same pair of states. For example, if we move a disk from peg A to peg B, and then another disk from peg C to peg D, applying these two moves in the opposite order will generate the same state.

Since it can't detect most duplicate nodes, a depth-first search must generate every path to a given depth. The number of such paths is the branching factor raised to the power of the depth of the search. For example, the optimal solution depth for the 6-disk 4-peg Towers of Hanoi problem is 17 moves. Thus, a brute-force depth-first search to this depth would generate about $3.7664^{17} \approx 6 \times 10^9$ nodes, making it the largest 4-peg problem that is feasibly solvable by depth-first search. By storing all generated nodes, however, a breadth-first search wouldn't expand more than the number of unique states in the problem space. Since any disk can be on any peg, and the disks on any peg must be in decreasing order of size, there are only $4^n$ states of the $n$-disk problem. For six disks, this is only $4^6 = 4096$ states. Thus, breadth-first search is much more efficient than depth-first search on this problem.

The heuristic analog of breadth-first search is the well-known A* algorithm (Hart et al., 1968). A* maintains a Closed list of expanded nodes, and an Open list of those nodes generated but not yet expanded. Since it stores every node it generates on one of these two lists, A* is limited by the amount of available memory, typically exhausting memory in a matter of minutes on current machines.

### 5.2.3 FRONTIER A* ALGORITHM

Frontier-A* (FA*) is a modification of A* designed to save memory (Korf, 1999; Korf & Zhang, 2000). Instead of saving both the Open and Closed lists, frontier-A* saves only the Open list, and deletes nodes from memory once they have been expanded. Its name derives from the fact that the Open list can be viewed as the frontier of the search space, whereas the Closed list corresponds to the interior of the space. In order to keep from regenerating closed nodes, with each node on the Open list the algorithm stores those operators that lead to closed nodes, and when expanding a node those operators are not applied. Once the goal is reached, we can't simply trace pointers back through the Closed list to construct the solution path, since the Closed list is not saved. Rather, a divide-and-conquer technique is used to reconstruct the solution path in frontier search. Since our purpose here is simply to compare different heuristic functions, in the experiments discussed below we only executed the first pass of the algorithm to reach the goal state and determine the solution depth, rather than reconstructing the solution path. For more details on frontier-A* and frontier search in general, the reader is referred to (Korf, 1999; Korf & Zhang, 2000).

## 5.3 Admissible Heuristic Evaluation Functions

Both A* and frontier-A* require a heuristic function $h(n)$ to estimate the distance from node $n$ to a goal. If $h(n)$ is admissible, or never overestimates actual cost, both algorithms are guaranteed to find an optimal path to the goal, if one exists.

### 5.3.1 HEURISTIC FUNCTION BASED ON INFINITE-PEG RELAXATION

The simplest admissible heuristic for this problem is the number of disks not on the goal peg. A more effective admissible heuristic comes from a relaxation of the problem which allows an infinite number of pegs, or equivalently, as many pegs as there are disks. In this relaxed problem, the disks on any non-goal peg can be moved to the goal peg by moving all but the bottom disk to their own temporary peg, moving the bottom disk to the goal peg, and then moving each of the remaining disks from their temporary pegs to the goal peg. Thus, the required number of moves of disks on each non-goal peg is twice the number of disks on the peg minus one. For disks on the goal peg, we count down from the largest disk until we find one that is not on the goal peg. Each smaller disk on the goal peg must move twice, once to its temporary peg to allow the largest disk not on the goal peg to be moved to the goal peg, and then from its temporary peg back to the goal peg. The number of moves needed to optimally solve this relaxed problem is the sum of these values for each peg. This is also an admissible heuristic for the original problem, which we call the *infinite-peg* heuristic.

### 5.3.2 PRECOMPUTING ADDITIVE PATTERN DATABASES

We can also compute additive pattern database heuristics for this problem. Consider a 15-disk problem, and assume we partition the disks into one group of 8 disks and another group of 7 disks. For example, we could divide them into the 8 largest disks and the 7 smallest disks, or any other disjoint partition we choose. Then, for each group of disks, we compute a table with one entry for every different possible combination of pegs that those disks could occupy. The value of the entry is the minimum number of moves required to move all the disks in the group to the goal peg, assuming there are no other disks in the problem. Since there are exactly $4^n$ states of the $n$-disk problem, indexing this table is particularly easy, since each disk position can be represented by two bits, and any configuration of $n$ disks can be uniquely represented by a binary number $2n$ bits long. We can compute such a table for each group of disks in a single breadth-first search backward from the goal state, in which all the disks in the group start on the goal peg, and there are no other disks. We store in the table the number of moves made the first time that each state is encountered in this breadth-first search.

There are two important simplifications we can take advantage of in this domain. The first is that for a given number of disks, the pattern database will contain exactly the same values regardless of which disks we choose. In other words, a pattern database for the eight largest disks, or the eight smallest disks, or eight even-numbered disks will all be identical. The reason is that only the number of disks matter, and not their absolute sizes. Therefore, we only need a single pattern database for each number of disks.

The second simplification is that a pattern database for $n$ disks also contains a pattern database for $m$ disks, if $m < n$. To look up a pattern of $m$ disks, we simply assign the $n - m$ largest disks to the goal peg, and then look up the resulting configuration in the $n$-disk pattern database. The result is that we only need a single pattern database for the largest number of disks of any group of our partition.

### 5.3.3 COMPUTING THE HEURISTIC VALUES FROM THE PATTERN DATABASE

The most effective heuristic is based on partitioning the disks into the largest groups possible, and thus we want the largest pattern database that will fit in memory. For example, a 14-disk database occupies $4^{14}$ entries or 256 megabytes, since the maximum possible value in such a database is 113 moves, which fits in one byte. This is the largest pattern database we can use on a machine with a gigabyte of memory, since a 15-disk database would occupy all the memory of the machine at one byte per entry. The 14-disk database gives us the exact optimal solution cost for any problem with 14 or fewer disks. To solve the 15-disk problem, we can partition the disks into 14 and 1, 13 and 2, 12 and 3, etc.

Given a particular split, such as 12 and 3, the static partitioning approach will always group the same 12 disks together, say for example, the 12 largest disks vs. the 3 smallest disks. Given a particular state of the search, we use the positions of the 12 largest disks as an index into the database, and retrieve the value stored there. We then look up the position of the 3 smallest disks, assigning all larger disks to the goal peg, and retrieve that value. Finally, we add these two values for the final heuristic value.

For any given state, which disks we choose to group together will affect the heuristic value. For example, consider a state of the 15-disk problem where the 3 largest disks are on the goal peg. If we group the 12 largest disks together, we won't get an exact heuristic value, because we won't capture the interactions with the 3 smallest disks. If instead we group the 12 smallest disks together, and put the 3 largest disks in the other group, our heuristic value will be exact, because no moves are required of the 3 largest disks, and they don't interact with any of the smaller disks once they are on the goal peg. As another example, if we group the 3 smallest disks together and they all happen to be on the goal peg, then no moves will be counted for them, whereas if they were grouped with at least one larger disk not on the goal peg, the heuristic will add at least two moves for each of the smaller disks on the goal peg.

The dynamic partitioning approach potentially uses a different partition of the disks for each state. Given a 12-3 split for example, we consider all $15 * 14 * 13/6 = 455$ ways of dividing the disks, compute the resulting heuristic value for each, and then use the maximum of these heuristic values.

| Heuristic | h(s) | Avg h | Nodes Expanded | Seconds | Nodes Stored |
|-----------|------|-------|----------------|---------|--------------|
| Infinite Peg | 29 | 26.37 | 371,556,049 | 1198 | 27,590,243 |
| Static 12-3 | 86 | 62.63 | 9,212,163 | 32 | 1,778,813 |
| Static 13-2 | 100 | 74.23 | 2,205,206 | 7 | 525,684 |
| Static 14-1 | 114 | 87.79 | 158,639 | 2 | 82,550 |
| Dynamic 14-1 | 114 | 95.52 | 122,128 | 2 | 25,842 |

Table 4: Results for the 15-disk problem

## 5.4 Experimental Results

For each of the experiments below, we used the frontier-A* algorithm, since A* ran out of memory on the larger problems or with the weaker heuristics. The database we used was the complete 14-disk pattern database, which requires 256 megabytes of memory. We report experiments with the 15, 16, and 17 disk problems.

### 5.4.1 15-Disk Problem

For the 15-disk problem, we implemented five different heuristic functions. The first was the infinite-peg heuristic, described in section 5.3.1 above. This heuristic doesn't require a database.

The next three heuristics were statically-partitioned heuristics, based on the 14-disk pattern database. We divided the disks into groups of 12 and 3, 13 and 2, and 14 and 1, respectively. For the 12-3 split, the tiles were divided into the 12 largest disks and the three smallest disks, and similarly for the 13-2 and 14-1 splits.

Finally, we used a dynamically-partitioned pattern database heuristic, dividing the disks into a group of 14 and a singleton disk. For each state, we computed the heuristic value for all 15 different ways of dividing the disks into groups of 14 and one, and took the maximum of these as the heuristic value.

In each case we solved the single problem instance in which all 15 disks start on a single peg, and the task is to transfer them all to a different goal peg. Ties among nodes of equal $f(n)$ value were broken in favor of smaller $h(n)$ values. The optimal solution path is 129 moves long. Table 4 shows the results. The first column lists the heuristic function used. The second column gives the heuristic value of the initial state. The third column shows the average value of the heuristic function over the entire problem space. The fourth column indicates the number of nodes expanded, and the fifth column presents the time in seconds, on a 450 megahertz Sun Ultra-10 workstation. The last column shows the maximum number of nodes stored, which is an indication of the relative memory required by the algorithms, excluding the memory for the pattern database.

The data show that the heuristic based on the infinite-peg relaxation is the weakest one, requiring the most time and storing the most nodes. It is also the slowest to compute per node, since the position of each disk must be considered in turn. In contrast, the database heuristics can be calculated by using the bits of the state representation as an index into the database, with shifting and masking operations to partition the bits.

| Heuristic | h(s) | Avg h | Nodes Expanded | Seconds | Nodes Stored |
|-----------|------|-------|----------------|---------|--------------|
| Static 13-3 | 102 | 75.78 | 65,472,582 | 221 | 10,230,261 |
| Static 14-2 | 114 | 89.10 | 17,737,145 | 65 | 2,842,572 |
| Dynamic 14-2 | 114 | 95.52 | 6,242,949 | 96 | 1,018,132 |

Table 5: Results for the 16-disk problem

Among the statically-partitioned heuristics, the data clearly show that the larger the groups, the better the performance. In particular, splitting the 15 tiles into 14 and 1 is much more effective than the 13-2 or 12-3 split.

Finally, the dynamically-partitioned 14-1 split generated the smallest number of nodes and thus consumed the smallest amount of memory. This is because it computes the best partition of the disks for each state, in order to maximize the heuristic value. While this heuristic takes longer to compute for each node, it is not a significant factor in the overall time, and it makes up for it by expanding and storing fewer nodes.

### 5.4.2 16-Disk Problem

For the 16-disk problem, the optimal solution length is 161 moves. Without a pattern database, we could store approximately 34 million nodes on our one-gigabyte machine. Frontier-A* with the infinite-peg heuristic ran out of memory while expanding nodes whose $f = g + h$ cost was 133.

Storing the 14-disk pattern database, which takes 256 megabytes of memory reduced the number of nodes we could store to about 23 million. This only allowed us to solve the problem with a 13-3 or 14-2 split. We compared heuristics based on statically partitioned splits of 13-3 and 14-2 against a dynamically-partitioned 14-2 heuristic. In the later case, there are $16 * 15/2 = 120$ ways to partition the 16 disks into one group of 14 disks and another group of 2 disks. For each state, we computed the heuristic for each different partition, and used the maximum heuristic value.

The results are shown in Table 5, in a format identical to that in Table 4 for the 15-disk problem. Again, we find that the 14-2 static split is much more effective than the 13-3 split. In this case, the dynamically-partitioned heuristic took longer to solve the problem than the largest static partitioning, since the dynamically-partitioned heuristic is much more expensive to compute than the statically-partitioned heuristic. On the other hand, the more accurate heuristic resulted in expanding and storing fewer nodes, thus requiring less memory, which is the limiting resource for best-first search algorithms such as A* or frontier-A*.

### 5.4.3 17-Disk Problem

Finally, we tested our heuristics on the 17-disk problem. The optimal solution length to this problem is 193 moves. We were able to store only about 23 million nodes without a pattern database, since the state representation requires more than a single 32-bit word in this case. Frontier-A* with the infinite-peg heuristic ran out of memory while expanding nodes of cost 131. With the 14-disk pattern database we were only able to store about 15 million nodes. With static partitioning of the disks into the largest 14 and the smallest 3, frontier-A* exhausted memory while expanding nodes of cost 183. The only way we were able to solve this problem was with a dynamically-partitioned 14-3 heuristic. In this case, for each state we computed all $17 * 16 * 15/6 = 680$ different ways of dividing the disks, looked-up the database value for each, and returned the maximum of these as the final heuristic value. This took two hours to run, expanded 101,052,900 nodes, and stored 10,951,608 nodes.

We conclude that with a gigabyte of memory, the dynamically-partitioned databases are better than the statically-partitioned databases for the 17-disk problem. If we had enough memory to solve a problem with both the statically- and dynamically-partitioned databases, using the statically partitioned database might be faster. However, the limiting resource for best-first searches such as A* is memory rather than time. Since the dynamically-partitioned databases use much less memory than the statically-partitioned databases, dynamic partitioning is the method of choice for solving the largest problems.

## 5.5 Time to Compute the Pattern Database

The results above do not include the time to compute the pattern database. Computing the 14-disk database takes about 263 seconds, or 4 minutes and 20 seconds. If we are solving multiple problem instances with the same goal state, but different initial states, then this cost can be amortized over the multiple problem solving trials, and becomes less significant as the number of problem instances increases.

The Towers of Hanoi problem traditionally uses a single initial state with all disks starting on the same peg, however. If that is the only state we care to solve, then the cost of generating the pattern database must be taken into account. For the 15-disk problem, this would add 263 seconds to each of the running times for the database heuristics in Table 4. Even with this additional time cost, they all outperform the infinite-peg heuristic, by as much as a factor of 4.5. The 16- and 17-disk problems are too big to be solved on our one gigabyte machine without the pattern database.

## 5.6 Capitalizing on the Symmetry of the Space

Frontier-A* with dynamically-partitioned pattern-database heuristics is the best existing technique for solving arbitrary initial states of large 4-peg Towers-of-Hanoi problems. If we are only interested in solving the standard initial state with all disks on a single peg, however, we can do better by capitalizing on the fact that the initial and goal states are completely symmetric in this case (Bode & Hinz, 1999).

Consider moving all $n$ disks from the initial to the goal peg. In order to be able to move the largest disk to the goal peg, all the $n-1$ smaller disks must be removed from the initial peg, cannot be on the goal peg, and hence must be on the remaining two intermediate pegs. Note that we don't know the optimal distribution of the $n-1$ smallest disks on the intermediate pegs, but we do know that such a state must exist on any solution path.

Once we reach a state where the largest disk can be moved to the goal peg, we make the move. If we then apply the moves we made to reach this state from the initial state in reverse order, and interchange the initial and goal pegs, instead of moving the $n-1$ smallest disks from the intermediate pegs back to the initial peg, we will move them to the goal peg, thus completing the solution of the problem. Thus, if $k$ is the minimum number of moves needed to reach a state where the largest disk can be moved to the goal peg, the optimal solution to the problem requires $2k + 1$ moves.

To find the optimal solution to the standard initial state, we simply perform a breadth-first search, searching for the first state where the largest disk can be moved to the goal peg. This only requires us to search to just less than half the optimal solution depth. Unfortunately, since we can't completely determine the goal state of this search, it is difficult to design an effective heuristic function for it. Thus, we used breadth-first frontier search.

In general, this technique is faster than our best heuristic search for the goal state, but requires storing more nodes. For example, for the 15-disk problem, the breadth-first search to the half-way point requires only 18 seconds, compared to 265 seconds for the heuristic search on a 450 megahertz Sun workstation, counting the time to generate the pattern database. On the other hand, it requires storing about 1.47 million nodes, compared to about 26,000 nodes for the heuristic search. For the 16-disk problem, the breadth-first search technique requires 102 seconds, but stores about 5.7 million nodes, compared to one million for the heuristic search. The 17-disk problem was the largest we could solve in one gigabyte of memory. It required 524 seconds and stored about 20 million nodes. By using disk storage for memory in a more complex algorithm (Korf, 2004), we have been able to solve up to 24 disks, in each case verifying the conjectured optimal solution.

Note that unlike the heuristic search, this technique only applies to the standard initial and goal states, where all disks are on one peg, and not to arbitrary problem instances.

Recently (Felner et al., 2004), we were able to significantly improve these results by compressing large pattern databases to a smaller size. We solved the 17-disk problem in 7 seconds and the 18-disk problem in 7 minutes, not counting the time to compute the pattern database.

## 6. Vertex Cover

We now turn our attention to our final domain, that of computing an optimal vertex cover of a graph. We begin with a description of the problem, then consider the problem space to be searched, and finally the computation of additive pattern databases heuristics for this domain. Note that we have already seen an application of a more general version of this problem, the weighted vertex cover, to the problem of computing an admissible heuristic for the sliding-tile puzzle.

### 6.1 The Problem

Given a graph, a vertex cover is a subset of the vertices such that every edge of the graph is incident to at least one of the vertices in the subset. A minimum or optimal vertex cover is one with the fewest number of vertices possible. Vertex cover was one of the earliest problems shown to be NP-complete (Karp, 1972). There are many approaches to solve this problem. A comprehensive survey is beyond the scope of this paper. We refer the reader to a survey that was presented by Balasubramanian et al. (Balasubramanian, Fellows, & Raman, 1998).

## 6.2 The Search Space

To avoid confusing the search space tree with the original input graph, we will use the terms "nodes" and "operators" to refer to parts of the search tree, and the terms "vertices" and "edges" to refer to parts of the input graph.

One of the simplest and most effective search spaces for this problem is a binary tree. Each node of the tree corresponds to a different vertex of the graph. The left branch includes the corresponding vertex in the vertex cover, and the right branch excludes the vertex from the vertex cover. The complete tree will have depth $n$, where $n$ is the number of vertices, and $2^n$ leaf nodes, each corresponding to a different subset of the vertices. (Downey & Fellows, 1995) classify this method as the *bounded search tree* method.

Of course, much of this tree can be pruned in the course of the search. For example:

- If a vertex is excluded from the vertex cover, every adjacent vertex must be included, in order to cover the edge between them. In such cases, branches that do not include these adjacent vertices are pruned.

- If a vertex is included, all of its incident edges are covered, so we modify the graph by removing this vertex and those edges for the remainder of the search below that node.

- If a vertex has no edges incident to it, we can exclude that vertex from the vertex cover.

- If a vertex has only one neighbor, we can exclude that vertex and include its neighbor. If there is an edge between two vertices that have no other neighbors, we can arbitrarily choose one of the vertices to include, and exclude the other.

Many attempts have been made recently to reduce the size of the search tree by pruning irrelevant nodes or by combining several nodes together (Chen, Kanj, & Jia, 2001; Balasubramanian et al., 1998; Downey, Fellows, & Stege, 1999; Niedermeier & Rossmanith, 1999). To date, the best of these methods is the work done by (Chen et al., 2001). However, despite the comprehensive and deep work that they have done, they all use a simple $g(n)$ cost function for each internal node $n$ of the search tree which is the size of the partial vertex cover associated with node $n$. We extend this work by adding an admissible heuristic function $h(n)$, based on additive pattern databases.

During the course of the search, we keep the vertices of the remaining graph sorted in decreasing order of degree, and always branch next on the inclusion or exclusion of a remaining vertex of maximum degree.

The best way to search this tree is using depth-first branch-and-bound (DFBnB). The inclusion-exclusion scheme guarantees that there is a unique path to each vertex cover, and hence no duplicate nodes, eliminating the need for storing previously generated nodes and allowing a linear-space depth-first search. Once we reach the first complete vertex cover, we store the number of vertices it includes. At any node $n$ of the tree, let $g(n)$ represent the number of vertices that have been included so far at node $n$. This is the cost of the partial vertex cover computed so far, and can never decrease as we go down the search tree. Thus, if we encounter a node $n$ for which $g(n)$ is greater than or equal to the smallest complete vertex cover found so far, we can prune that node from further search, since it cannot lead to a better solution. If we find a complete vertex cover with fewer nodes than the best so far, we update the best vertex cover found so far.

## 6.3 Heuristic Evaluation Functions

Our main contribution to this problem is adding a heuristic evaluation function to estimate the number of additional vertices that must be included to cover the edges of the remaining graph. Denote such a function $h(n)$, where $n$ is a node of the search tree. In our DFBnB algorithm, at each interior node $n$ of the search, we apply the cost function $f(n) = g(n) + h(n)$. If this cost equals or exceeds the size of the smallest vertex cover found so far, we can prune this node. We are still guaranteed an optimal solution as long as $h(n)$ never overestimates the number of additional vertices we have to add to the current vertex cover.

For example, consider two disjoint edges in the remaining graph, meaning they don't share either of their end points. At least two vertices must be added to the vertex cover, since one vertex incident to each edge must be included, and no vertex is incident to both edges. More generally, a matching of a graph is a set of vertex-disjoint edges, and a maximum matching is a matching with the largest number of such edges. A maximum matching can be computed in polynomial time (Papadimitriou & Steiglitz, 1982). Given a remaining graph, any vertex cover of it must be at least as large as a maximum matching of the graph. Thus, the size of a maximum matching is an admissible heuristic function for the vertex-cover problem.

The size of a maximum matching is not the best admissible heuristic, however, as we can consider costs of larger groups of vertices. For example, consider a triangle graph of three nodes with an edge between each pair, as shown in Figure 2. A maximum matching of this graph contains one edge, since any two edges have a vertex in common. A minimum vertex cover of the triangle contains two vertices, however, since no single vertex is incident to all three edges.

More generally, consider a clique of $k$ vertices, which is a set of vertices with an edge between every pair. A minimum vertex cover of such a clique includes $k-1$ vertices. This is sufficient, since the edges incident to the single vertex left out are covered by the remaining included vertices. $k-1$ vertices are necessary, because if any two vertices are excluded, the edge between them will not be covered. A single edge is the special case of a clique of size two. The idea of using cliques as a lower bound on the size of the minimum vertex cover is due to (Marzetta, 1998).

### 6.3.1 COMPUTING AN ADMISSIBLE HEURISTIC WITH ADDITIVE PATTERN DATABASES

In order to compute an admissible heuristic for the size of the minimum vertex cover of a given graph, we partition the vertices into a set of vertex-disjoint cliques, meaning two cliques can't share a vertex. Each clique of size $k$ contributes $k-1$ vertices to the size of the vertex cover, and we can sum these values for a set of vertex-disjoint cliques. In general the largest cliques make the largest contribution to the heuristic value. We would like the partition that yields the maximum value.

The *clique* problem is known to be NP-complete. However, we have found that for the random graphs that we experimented with (with fixed average degree), cliques of four or more vertices are very rare. Thus, in the experiments described below, we only looked for cliques of four nodes or less in time $O(n^4)$.

Building an additive pattern database for this problem is done as follows. We first scan the original input graph and identify all cliques of four or fewer nodes. These cliques are then stored in a pattern database. Then, during the search we need to retrieve disjoint cliques from this database as will be shown below.

### 6.3.2 Computing the Heuristic from the Pattern Database

After building the cliques database for the original graph, we calculate the heuristic for a given remaining subgraph. A special case of such a graph is the complete input graph at the root of the search tree where none of the vertices are included yet.

We compute the heuristic as follows. We scan the cliques database and find those cliques all of whose vertices remain in the given graph. Note that each of these cliques is now a hyperedge in the mutual-cost graph. We then would like to partition these cliques into a set of vertex-disjoint cliques to maximize the total number of vertices included. For each such clique of size $k$, we add $k - 1$ to the heuristic value.

Unfortunately, this partitioning problem is NP-complete. To avoid this NP-complete problem, we use a greedy approach that looks for a *maximal* set of cliques. A *maximal* set of cliques is a set of vertex-disjoint cliques that cannot be increased by adding more cliques to it, whereas a *maximum* set of cliques is one for the which the total number of vertices is as large as possible. In practice, we use a greedy approach that first identifies vertex-disjoint cliques of size four, then cliques of size three, and finally additional edges.

### 6.3.3 Static Vs. Dynamic Partitioning

The above algorithm gives us a relatively efficient way to compute a lower-bound on the size of the minimum vertex cover for a given graph. During our branch-and-bound search, however, as vertices are included and excluded, the remaining graph is different at each node. We need to compute a heuristic value for each of these remaining graphs based on data from the cliques database.

There are at least two ways to do this, corresponding to a static partition of the vertices and a dynamic partition.

One way to compute a heuristic for each node is to partition the original graph into vertex-disjoint cliques in order to obtain a vertex cover of maximal size, and then use this same partition throughout the search. With this approach, at each node of the search, we examine the remaining graph. For each clique of size $k$ in our original partition, if $k - 1$ of the vertices of the partition have already been included, we don't add any to the heuristic, since all the edges of this clique have been covered. If fewer then $k - 1$ vertices of the clique have been included so far, we add the difference between $k - 1$ and the number of vertices included to the heuristic value, since at least this many additional vertices of this clique must eventually be included. We do this for all the cliques in the partition, and sum the resulting contributions to determine the heuristic value of the node. We refer to this as the statically-partitioned heuristic.

| Heuristic | Cliques | Nodes | seconds | nodes/sec |
|---|---|---|---|---|
| No heuristic | – | 720,582,454 | 4,134 | 174,306 |
| Static | 2 | 187,467,358 | 1,463 | 128,139 |
| Static | 3 | 105,669,961 | 849 | 124,464 |
| Static | 4 | 103,655,233 | 844 | 123,252 |
| Dynamic | 2 | 19,159,780 | 231 | 82,982 |
| Dynamic | 3 | 4,261,831 | 70 | 60,451 |
| Dynamic | 4 | 4,170,981 | 73 | 57,136 |

Table 6: Performance of the different algorithms on graphs with 150 vertices and a density of 16.

Dynamic partitioning is done by repartitioning the vertices of the remaining graph into vertex-disjoint cliques at each node of the search tree, and computing the resulting heuristic. This is done according to the methods described in section 6.3.2.

Clearly, the dynamically-partitioned heuristic will be more accurate than the statically-partitioned heuristic, since the dynamic partition is based on the current remaining graph at each node. On the other hand, the dynamically-partitioned heuristic will be more expensive to compute, since we have to find a maximal partitioning of the remaining graph at each node. To compute the statically-partitioned heuristic, we only have to do the partitioning once, and then during the search we simply examine each node in each clique of the original partition to determine if it has been included or not.

### 6.4 Experiments

We compared depth-first branch-and-bound with the statically-partitioned heuristic to depth-first branch-and-bound with the dynamically-partition heuristic. We experimented on random graphs which were built as follows. Given two parameters, $N$ to denote the number of vertices and $D$ to denote the density (average degree of the graph) , we created a graph with $N$ vertices such that each edge was added to the graph with probability $\frac{D}{N}$. Table 6 presents the results of solving the vertex-cover problem on random graphs with 150 vertices and density of 16. The *Cliques* column presents the size of the largest clique that we store in our database. The remaining columns present the number of generated nodes and the number of seconds needed to solve an average problem. All the experiments reported here were also tried on other densities and the same tendencies were observed. We focus on density of 16 because it seems to be the most difficult case.

For the vertex cover problem each problem instance has its own database. However, in all our experiments it took only a fraction of a second to build the database and hence this time is omitted in all our data.

As expected, the dynamically-partitioned heuristic generated fewer nodes than the statically-partitioned heuristic, but ran slower per node. Overall, however, the dynamically-partitioned heuristic outperformed the statically-partitioned heuristic by up to an order of magnitude in running time. Both heuristics clearly outperform a system where no heuristics were used. We have tried our system on many other graphs and obtained similar results.

We have also added our heuristics to the best proven tree search algorithm for vertex cover, which was proposed by (Chen et al., 2001). They divide the nodes of the search tree into a number of classes. Each class is determined by looking at special structures of vertices (like the degree of the vertex) and edges of the input graph in the area of the relevant vertex. Due to the special structure of the graph, some decisions on these vertices are forced and thus the size of the search tree is reduced by pruning other possibilities for branching. For example, consider a degree-2 vertex $v$ whose two neighbors $u$ and $w$ are connected to each other. In that case, since the three vertices form a triangle, $u$ and $w$ can be added to a minimal vertex-cover. (Chen et al., 2001) provided a large number of different classes and cases for pruning the search tree by focusing on special classes of vertices or by combining several vertices together. They prove that their algorithm is asymptotically the fastest algorithm that optimally solves the vertex-cover problem. Chen's algorithm calculates the cost of a node as the size of the partial vertex cover and does not try to gather knowledge from the vertices in the remaining graph in order to get a lower bound estimation on the number of vertices that must be added to the partial vertex cover. In other words they only use a cost function of $f = g$. We have added our heuristics to their system and use the cost function of $f = g + h$. Adding the dynamic database described above to Chen's pruning methods further improved the running time as will be shown below.

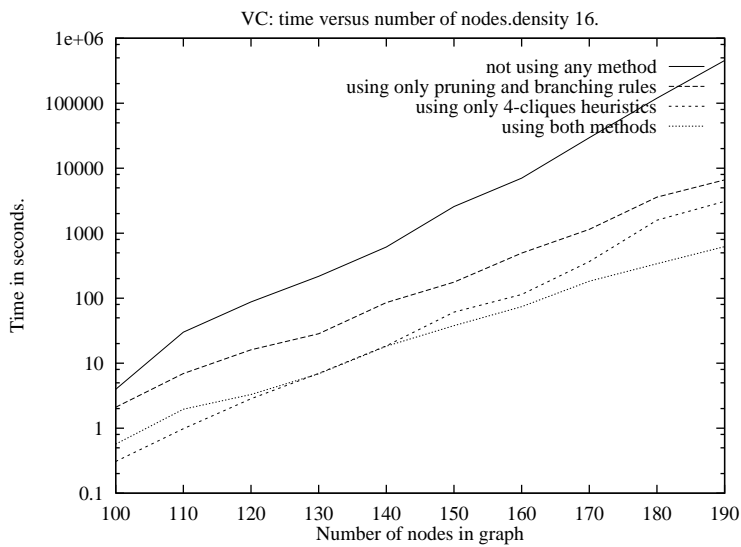

Figure 7: Combining both methods

Figure 7 shows four curves that include all the four ways to combine the 4-cliques heuristics with the pruning methods proposed by (Chen et al., 2001). The four ways are: use them both, use only one of them or not using any of them. The figure presents time in seconds on random graphs with sizes of 100 up to 190 and density of 16. The results clearly show that when using each method separately, the heuristic function approach outperforms the pruning methods. Figure 7 also shows that when the input graph is larger than 140, the best combination between the two approaches is to use both of them. In other words, adding our heuristics to the algorithm of (Chen et al., 2001) further improves the running time. On graphs with less than 140 nodes, however, it is even preferable to use the most accurate heuristic function without any pruning method. In all cases, adding our 4-cliques pattern database heuristic to any combination of the pruning methods is always beneficial. For example, for a graph with 190 vertices, the simple system with no pruning methods and no heuristics needed 453,488 seconds. Adding our 4-cliques heuristics improved the running time to only 2,764 seconds. Adding the heuristics to a system with all the pruning methods further improved the overall time from 7,131 seconds for the pure Chen algorithm to 767 seconds for the combined algorithm.

| size | Random graph, density 8 | | | Random graph, density 16 | | | Delaunay graphs | | |
|---|---|---|---|---|---|---|---|---|---|
| Size | Vc | Nodes | Sec | Vc | Nodes | Sec | Vc | Nodes | Sec |
| 150 | 92 | 2400 | 1 | 113 | 316,746 | 26 | 103 | 27,944 | 1 |
| 200 | 130 | 296,097 | 33 | 153 | 21,897,066 | 2,045 | 137 | 224,349 | 3 |
| 250 | 164 | 9,703,639 | 1,116 | 187 | 544,888,130 | 58,776 | 171 | 4,035,989 | 67 |
| 300 | 181 | 56,854,403 | 7,815 | | | | 204 | 32,757,219 | 493 |
| 350 | 219 | 137,492,886 | 19,555 | | | | 241 | 1,146,402,687 | 20,285 |
| 400 | | | | | | | 269 | 27,443,208,087 | 485,581 |

Table 7: Experimental results of our best combination on different graphs.

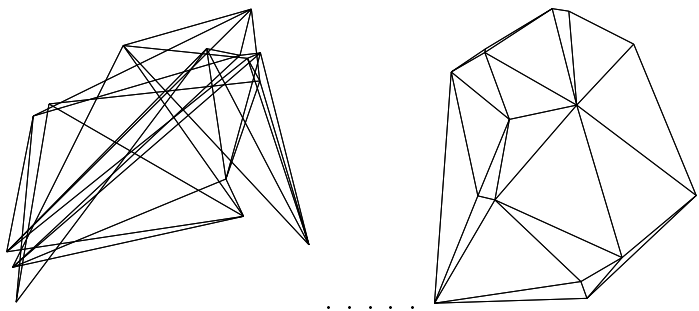

Figure 8: Random and Delaunay graphs of size 15.

Since combining both methods seems to be best, we used this combination to solve random graphs of larger sizes with densities of 8 and 16. We have also experimented with Delaunay graphs. These graphs comprise Delaunay triangulations of planar point patterns that are constructed by creating a line segment between each pair of points *(u,v)* such that there exists a circle passing through $u$ and $v$ that encloses no other point. Figure 8 illustrate a random graph and a Delaunay graph of size 15. In a random graph a vertex can be connected to any of the other vertices while in a Delaunay graph vertices are connected only to nearby vertices. Note that the average degree of Delaunay graphs is around 5.

The results are presented in Table 7. A graph of 250 vertices and a density of 16 was solved in 16 hours while with density of 8 we could solve a graph of 350 vertices in 5 hours. We could solve Delaunay graphs of size up to 400 vertices.

## 7. General Characterization of the Method

Here we abstract from our three example problems a general characterization of additive heuristic functions and pattern databases. While the sliding-tile puzzles and Towers of Hanoi are similar *permutation problems*, the vertex-cover problem is significantly different.

We begin by noting that additive heuristics and pattern databases are separate and orthogonal ideas. Additive heuristics are added together to produce an admissible heuristic. Pattern databases are a way of implementing heuristic functions by precomputing their values and storing them in a lookup table. These two techniques can be used separately or in combination. For example, Manhattan distance is an additive heuristic that is often computed as a function, rather than stored in a pattern database. Existing pattern database heuristics for Rubik's Cube are not additive. The heuristics described earlier for the sliding-tile puzzles and Towers of Hanoi are both additive and implemented by pattern databases. A heuristic such as Euclidean distance for road navigation is neither additive nor implemented by a pattern database. We begin by giving a precise characterization of a problem and its subproblems, then consider additive heuristics, and finally consider pattern databases.

### 7.1 Problem Characterization

A *problem space* is usually described abstractly as a set of atomic states, and a set of operators that map states to states. This corresponds to a labeled graph, called the problem-space graph. In addition, a specific problem instance is a problem space together with a particular initial state and a (set of) goal state(s). In order to talk precisely about subproblems, however, we need to add more structure to this abstract characterization. For this, we adopt a formalism used by Korf (Korf, 1985b), and by others as well.

#### 7.1.1 PROBLEM STATE

We define a problem *state* as a vector of state variables, each of which is assigned a particular value. For example, in the Towers of Hanoi problem, there is a variable for each disk, with a value that indicates which peg the disk is on. For the sliding-tile puzzles, there is a variable for each physical tile, plus the blank position, with a value that indicates the position occupied by that tile or the blank. We represent the blank position explicitly to easily determine if a particular move is applicable. For the vertex-cover problem, there is a variable for each edge of the graph, with a value of one or zero, indicating whether the edge has been covered by an included vertex or not. For ease of exposition, below we will often refer to the variable that corresponds to a particular disk, tile, or edge as simply that disk, tile, or edge.

The initial state is a particular set of values assigned to each variable in the state vector. For the permutation problems, this specifies the initial positions of the tiles and blank, or disks. For the vertex-cover problem, the initial state assigns zero to each edge, indicating that none of them are covered yet. A goal state is also indicated by a set of assignments to the variables. For the permutation problems this reflects the desired positions of the tiles or disks. For the vertex-cover problem, the goal state assigns one to each edge, indicating that each edge is covered.

### 7.1.2 OPERATORS

An operator in this formulation is a partial function from a state vector to a state vector, which may or may not apply to a particular state. For the vertex-cover problem, there is an operator for each vertex. To include a vertex in the solution, we apply the corresponding function, which sets the value of each edge incident to the vertex to one, regardless of its previous value, and leaves the values of the other edges unchanged. For the sliding-tile puzzles, we define a separate operator for each physical tile. The corresponding function applies to any state where the blank is adjacent to the given tile, and swaps the positions or values of the blank variable and the tile variable. For the Towers of Hanoi problem, there is a separate operator for each combination of disk and destination peg. If none of the smaller disks are on the same peg as the disk to move, nor the destination peg, then the effect of the operator is to change the value of the moving disk to the destination peg. If this condition doesn't hold, then the operator is not applicable. Note that there are other natural ways to define the set of operators for the sliding-tile puzzles or the Towers of Hanoi.

We say that an operator function application *affects* a variable if it changes its value. For the Towers of Hanoi, each operator only affects a single variable, that of the disk being moved, even though its applicability depends on all smaller disks. For the vertex-cover problem, each operator (vertex) affects all the variables (edges) that it is incident to. For the sliding-tile puzzle, each operator affects two variables, the physical tile being moved and the blank variable.

### 7.1.3 SOLUTIONS

A feasible solution to a problem is a sequence of operator function applications that maps the initial state to a goal state. The cost of a solution is the sum of the costs of the operators applied, and an optimal solution is one of lowest cost. There may be multiple optimal solutions. For example, in the vertex-cover problem, the order in which vertices are included doesn't matter, so an optimal set of operators can be applied in any order. In each of these problems, each operator costs one unit, but in general different operators could have different costs.

### 7.1.4 PATTERNS AND THEIR SOLUTIONS

We define a *pattern* as an assignment of values to a subset of the state variables in a problem. For example, a pattern in a permutation problem might be a particular configuration of a subset of the tiles or disks. A pattern is thus an abstract state that corresponds to a set of complete states of the problem which we call the *pattern set*. All the states in the pattern set have the same values assigned to the subset of variables included in the pattern but have different values for other variables.

Given a solution to a problem instance, we define the cost of solving a pattern in that problem instance as the sum of the costs of those operators in the solution that affect at least one variable included in the pattern. This is a key difference between our approach and that of (Culberson & Schaeffer, 1998), since they count the cost of all operators applied to solve a pattern, including those operators that only affect variables that are not included in the pattern. Even though other operators must often be applied to establish preconditions for operators that affect the variables in the pattern, we don't count them in the cost of the solution of the pattern. The optimal cost of solving a pattern is the lowest cost of solving the pattern from any complete state in the original problem in which the pattern variables have their initial values, or in other words, from any state that is a member of the pattern set. A solution of a pattern is therefore a lower bound on the solution to the complete state in the original problem. This is true for both our definition of solution to a pattern, and for the definition of (Culberson & Schaeffer, 1998).

## 7.2 Additive Heuristic Functions

The central question of this section is when is the sum of the optimal costs of solving a collection of patterns a lower bound on the optimal cost of solving the original problem? A sufficient condition is given below.

**Theorem 1** *If we partition a subset of the state variables in a problem instance into a collection of subsets, so that no operator function affects variables in more than one subset, then the sum of the optimal costs of solving the patterns corresponding to the initial values of the variables in each subset is a lower bound on the optimal cost of solving the original problem instance.*

Before giving the proof of this result, we illustrate it with each of our example domains. In the Towers of Hanoi problem, if we partition the disks into groups that are disjoint, the sum of the costs of solving the corresponding patterns will be an admissible heuristic, because each operator only affects one disk. In the vertex-cover problem, we have to partition the edges into groups so that no vertex is incident to edges in more than one group, because each vertex represents an operator that affects all the edges that are incident to it. This is most easily done by partitioning the vertices into disjoint subsets, and including in each subset only the edges between vertices in the same subset. In a connected graph, this will require leaving some edges out of all patterns. The cost of covering these excluded edges may not be counted, but that doesn't affect the admissibility of the heuristic. In the sliding-tile puzzles, we need to partition the physical tiles into disjoint subsets. We can't include the blank in any of these subsets, because every operator affects the blank position.

The proof of this result is very simple, now that we've made the right definitions.

**Proof:** Let $S$ be an optimal or lowest-cost solution of a complete problem instance. $S$ consists of a sequence of operators that map all the variables in the problem instance from their initial values to their goal values. The cost of $S$ is the sum of the costs of the individual operators in $S$.

$S$ is also a solution to any pattern in the original problem instance, since it maps any subset of the variables from their initial values to their goal values. Thus, it is a solution to each of the patterns in our hypothesized partition. Given a particular pattern, the cost of $S$ attributed to the pattern is the sum of the costs of those operators in $S$ that affect any of the variables in the pattern. Since by assumption no operator affects variables in more than one pattern in our partition, each operator of $S$ can be attributed to at most one pattern, based on the variables it affects. The cost of $S$ is thus greater than or equal to the sum of the costs of $S$ attributed to each of the patterns. Note that some variables may not be included in any patterns, and thus the costs of any operators that only affect such variables will not be attributed to the solution of any included patterns.

Since the optimal cost of solving a pattern is less than or equal to the cost of any solution to the pattern, we can conclude that the sum of the optimal costs of solving these patterns is a lower bound on the cost of the optimal solution to the original problem. Q.E.D.

Rubik's cube is another example of a permutation problem. In Rubik's cube, however, each operator moves multiple physical subcubes. Furthermore, there is no non-trivial way to partition the subcubes into disjoint subsets so that the operators only affect subcubes from the same subset. Thus, there are no known additive heuristics for Rubik's cube, and the best way to combine costs of different subproblems is to take their maximum. Similarly, the original pattern databases of (Culberson & Schaeffer, 1998), which counted all moves required to position the pattern tiles, including moves of non-pattern tiles, do not result in additive costs, even if the pattern tiles are disjoint.

An important question is how to divide the variables in a problem into disjoint patterns. There are two considerations here. One is that we would like a partition that will yield a large heuristic value. This favors partitioning the problem into a small number of patterns, each of which include a large number of variables. The other consideration is that we need to be able to efficiently compute, or precompute and store, the costs of solving the corresponding patterns. In general, the fewer the variables in a pattern, the easier it is to compute and store its solution cost, but the sum of such solutions tends to be a weaker lower bound on the overall solution cost, since the interactions between the patterns are not included in the heuristic.

For vertex-cover, any partition of the graph into edge-disjoint subgraphs will yield an admissible heuristic. We chose to partition the graphs into subgraphs which are cliques, however, because the cost of a vertex cover of a clique of $n$ nodes is both large relative to the number of nodes, and easy to compute, since it is always equal to $(n - 1)$.

## 7.3 Pattern Databases

A pattern database is a particular implementation of a heuristic function as a lookup table. It trades space for time by precomputing and storing values of the heuristic function. This only makes sense if the same values will be needed more than once, and there is sufficient memory available to store them. The values stored are the costs of solving subproblems of the original problem.

A pattern database is built by running a breadth-first search backwards from the complete goal state, until at least one state representing each pattern is generated. The cost of the solution to the pattern is the sum of the costs of the operators that changed any of the variables in the pattern. For the Towers of Hanoi problem, we can ignore any variables that are not part of the pattern. For the sliding-tile puzzle, we can ignore the tile variables that are not part of the pattern, but must include the blank tile, to determine the applicability of the operators.

### 7.3.1 COMPLETE PATTERN DATABASES

In general, a pattern database corresponds to a subset of the variables of the problem. A complete pattern database stores the cost of solving every pattern, defined by the set of all possible combinations of values of the variables in the pattern subset. For example, in a permutation problem, a pattern database corresponds to a particular set of tiles or disks, and a complete such database would contain the cost of solving any permutation of those tiles or disks. In the vertex-cover problem, a pattern database would correspond to a subset of the edges of the original graph. A complete pattern database would contain the number of additional vertices that must be included to cover every possible subset of that subset of edges.

One advantage of a complete pattern database is that every value is stored. Another advantage is that the pattern itself can often be used to compute a unique index into the database, so that only the cost of solving it need be stored. The primary disadvantage of a complete database is its memory requirements.

### 7.3.2 Partial Pattern Databases

A partial pattern database only stores the cost of solving some of the patterns. This saves memory by storing fewer values than a complete database. For example, given a subset of sliding tiles, a partial pattern database might only store those permutations of the tiles whose solution cost is more than the sum of their Manhattan distances. For the remaining patterns, we would just compute the sum of the Manhattan distances. This turned out to be very efficient for the dynamically-partitioned database as many of the values for the complete database were equal to the sum of the Manhattan distances. Another example is that in the tile puzzles we only stored triples whose values are not captured by any of their internal pairs. One disadvantage of a partial database is that since a value isn't stored for every pattern, we have to explicitly store the patterns that are included, in addition to their solution costs.

For our vertex-cover pattern databases, we store cliques of the original graph, up to a certain maximum size. Technically, each of these cliques would represent a different database, because each represents a different subset of the edges. However, we refer to the entire collection as a single database. Furthermore, we don't have to store a separate cost for each pattern of a clique, since in a clique of $n$ vertices of which $k$ vertices are already included, the number of additional vertices needed to cover all the edges of the clique is simply $n - 1 - k$. Thus, all we really need to store is the vertices of the clique itself, since the rest of the information needed is readily computed.

## 7.4 Algorithm Schemas

Both the statically-partitioned and the dynamically-partitioned pattern database heuristics include two phases, the pre-computation phase and the search phase. The steps for each of these phases are presented below. For both versions we first have to verify that additivity applies according to Theorem 1.

### 7.4.1 The Steps for the Statically-Partitioned Database Heuristics

- In the pre-computation phase do the following:

  - Partition the variables of the problem into disjoint subsets.

  - For each subset of variables, solve all the patterns of these variables and store the costs in a database.

- In the search phase do:

    - for each state of the search space, retrieve the values of solving the relevant patterns for each subset of the variables from the relevant databases, and add them up for an admissible heuristic.

### 7.4.2 THE STEPS FOR THE DYNAMICALLY-PARTITIONED DATABASE HEURISTICS

.

- In the pre-computation phase do the following:

    - For each subset of variables to be used (i.e., pair of variables or triple of variables etc.)
    - Solve all the patterns of these subsets and store the costs in a database.

- In the search phase do:

    - For each state in the search, retrieve the costs of the current patterns for each subset of variables.
    - Find a set of disjoint variables from the database such that the sum of the corresponding costs is the largest admissible heuristic. In general, the corresponding hypergraph matching problem is NP-complete, and we may have to settle for an approximation to the largest admissible heuristic value.

There are many domain-dependent enhancements that can be applied to the above schemas. For example, the question of whether to store a complete pattern database or a partial pattern database is domain dependent. Also, in some cases we may be able to improve the heuristic for a particular problem. In the tile puzzles for example, we used the solution to a weighted vertex-cover problem as an admissible heuristic.

## 7.5 Differences Between the Applications

While there are many commonalities among the three applications that we study here, there are also some important differences.

For the sliding-tile puzzles and Towers of Hanoi problem, we have a goal state that is often shared by many different problem instances. In that case, we need to compute only one set of pattern databases for all problem instances that share the same goal state. For these problems the time to create these tables can be amortized over many problem instances and can be omitted.

For the vertex-cover problem, every problem instance needs its own set of pattern databases. Thus, the time to compute these tables must be counted in the solution cost. However, our experiments show that the relative amount of time to compute these tables is insignificant, compared to the total time to solve large problems.

The vertex-cover problem is somewhat different than the permutation problems. In the permutation problems, the costs of solving the different instances of the different subproblems were explicitly stored in the database. For the vertex-cover problem we only stored the fact that a given subset of vertices form a clique. This is because calculating the exact cost of covering a given clique for each search node can be done very easily on the fly. For a clique of size $k$, the cost is $k - 1 - m$ if $m$ vertices of the clique are already included. One might claim that this is a degenerate form of pattern databases since we do not store the costs with the databases.

To summarize, while we have presented a general schema, trying this on new problems may involve significant creativity in identifying the subproblems, determining the definition of disjoint subproblems, computing the pattern databases, doing the partitioning, etc. Also, taking advantage of domain-dependent attributes is often very important.

## 8. Discussion, Conclusions and Future Work

A natural question to ask is given a problem, which type of database to use. While we can't give a complete answer to this question, we can offer some guidelines based on our experiments.

### 8.1 When to use each of the methods

The benefit of static partitioning is that it is done only once. Thus, the time per node for computing the heuristic is very low, usually no more than retrieving and summing a few values from a table. Dynamic partitioning considers multiple groups and partitions them on the fly. This produces a more accurate heuristic at a cost of more time per node. This overhead comes from updating the necessary data structures to reflect the current costs of the different groups, and from computing a maximum-matching to return the largest admissible heuristic.

One can add more patterns to a dynamic database to improve the heuristic. It appears however, that this is beneficial only up to a certain point. Beyond that point, the overhead per node dominates the reduction in the number of generated nodes, and thus the overall time increases. For example, for the tile puzzles we observed that the best balance is to only use pairs and triples. When we added quadruples to this system, we obtained a better heuristic, but the overall time to solve a problem increased. We can observe exactly the same behavior from the last two lines of Table 6 for the vertex-cover problem. For the graphs of size 150, adding cliques of size four reduced the number of generated nodes, but increased the overall time from 70 seconds to 73 seconds. Note however, that this was not the case for other sizes of graphs that we tested. See (Felner, 2001) for more details. For the 4-peg towers of Hanoi problem we observed the same phenomenon. For the 15-disk problem, the 14-1 dynamic partition solves the problem at the same speed as the 14-1 static partitioning, since there are only 15 different candidates to choose from. For the 16-disk problem however, there are 120 different partitions with a 14-2 split. This causes a larger overhead, and the dynamic partitioning runs slower than the static partitioning.

The question is where is the point at which adding more patterns is no longer effective. In this paper we compared two relatively extreme cases of the method where in one case we only had one possible partition and in the other we had as many partitions as possible. The optimal way might be somewhere in the middle. At this point we believe that given a specific problem, only trial and error experimentation will reveal the best option.

Another issue is memory requirements. Depending on the problem domain, different techniques may require more or less memory than others. For example, in the sliding-tile puzzles, the most effective dynamically-partitioned databases occupied much less memory than the best statically-partitioned heuristics. For the Towers of Hanoi problem, the memory requirements were the same for either method.

Note that if we use A* for frontier-A* as the search algorithm, then there is a competition for memory between the pattern databases and the need to store nodes of the search tree in Open and/or Closed lists. In that case, there might be a benefit for using the dynamic approach if it provides a more accurate heuristic. With a more accurate heuristic, a smaller number of nodes will be generated and the memory requirements of the Open and Closed lists will be decreased. This will allow us to solve larger problems with a given amount of available memory. An example of this is the 4-peg 17-disk Tower of Hanoi problem, which we could only solve with the more accurated dynamically-partitioned pattern databases.

## 8.2 Conclusions and Further Work

We have considered both static and dynamic additive pattern database heuristics in three different domains: sliding-tile puzzles, the 4-peg Towers of Hanoi problem, and vertex cover. In each case, the resulting heuristics are the best known admissible heuristics for these problems, and heuristic search using these heuristics is the best known method of finding optimal solutions to arbitrary instances these problem. For the special case of the standard initial and goal states of the Towers of Hanoi problem, however, we can use the symmetry between the initial and goal states to solve larger problems with breadth-first search (Korf, 2004).

In this paper we have studied the similarities and the differences between the two versions of additive pattern databases. We conclude that the question of which version to use is domain dependent. For example, for the Fifteen and Twenty-Four puzzles, heuristics based on a static partition of the problem are most effective. For the Thirty-Five puzzle, dynamic partitioning may be more effective. For the Towers of Hanoi problem and vertex cover, dynamically partitioning the problem for each state of the search is most effective for large problems.

There are a number of avenues for future work on admissible heuristics for these problem domains. Most recently, we have focussed on a technique for compressing larger pattern databases (Felner et al., 2004). Another direction is automatically finding the best static partitioning. Another avenue is to find more efficient algorithms for finding the best dynamic partitionings, rather than simply computing them all and taking their maximum. Another issue is to automatically find the optimal size of dynamically-partitioned databases.

Finally, we believe that these techniques can be effectively extended to new problem domains. Hopefully the diversity of domains represented in this paper and the general discussion will help in finding new applications of these ideas.

## 9. Acknowledgments

The work was carried out while the first author was at Bar-Ilan University. We would like to thank Eitan Yarden and Moshe Malin for their help with the code for the dynamic databases for the tile puzzles. We would like to thank Ido Feldman and Ari Schmorak for their help with the code for the 4-peg Towers of Hanoi problem. Thanks to Robert Holte for some fruitful discussions and for suggesting the 4-peg Towers of Hanoi problem. Richard Korf's work was supported by NSF under grant No. EIA-0113313, by NASA and JPL under contract No. 1229784, and by the State of California MICRO grant No. 01-044.